\def\year{2022}\relax
\documentclass[letterpaper]{article} % DO NOT CHANGE THIS
\usepackage{aaai22}  % DO NOT CHANGE THIS
\usepackage{times}  % DO NOT CHANGE THIS
\usepackage{helvet}  % DO NOT CHANGE THIS
\usepackage{courier}  % DO NOT CHANGE THIS
\usepackage[hyphens]{url}  % DO NOT CHANGE THIS
\usepackage{graphicx} % DO NOT CHANGE THIS
\def\UrlFont{\rm}  % DO NOT CHANGE THIS
\usepackage{natbib}  % DO NOT CHANGE THIS AND DO NOT ADD ANY OPTIONS TO IT
\usepackage{caption} % DO NOT CHANGE THIS AND DO NOT ADD ANY OPTIONS TO IT
\DeclareCaptionStyle{ruled}{labelfont=normalfont,labelsep=colon,strut=off} % DO NOT CHANGE THIS
\usepackage{algorithm}
\usepackage{algorithmic}

\usepackage{newfloat}
\usepackage{listings}
\floatstyle{ruled}
\newfloat{listing}{tb}{lst}{}
\floatname{listing}{Listing}
\usepackage{amsmath}
\usepackage{amssymb}
\usepackage{multirow}
\usepackage{booktabs}
\usepackage{array}
\usepackage{bm}
\usepackage{bbm}
\usepackage[table]{xcolor}
\usepackage{placeins}

\usepackage{lipsum}
\newcommand\blfootnote[1]{%
\begingroup
\renewcommand\thefootnote{}\footnote{#1}%
\addtocounter{footnote}{-1}%
\endgroup
}

\usepackage[colorlinks,hyperindex,breaklinks]{hyperref}

\title{Meta-DETR: Image-Level Few-Shot Object Detection with \\ Inter-Class Correlation Exploitation}
\author{
    {\fontsize{10.0}{1cm}\selectfont  Gongjie Zhang$^\dagger$ \qquad Zhipeng Luo$^\dagger$ \qquad Kaiwen Cui \qquad Shijian Lu$^*$}
}
\affiliations{
    {Nanyang Technological University, Singapore}\\
    {\tt\scriptsize gongjiezhang@ntu.edu.sg \quad zhipeng001@e.ntu.edu.sg \quad kaiwen001@e.ntu.edu.sg \quad shijian.lu@ntu.edu.sg}
}

\usepackage{bibentry}
\begin{document}

\maketitle

\begin{abstract}

Few-shot object detection has been extensively investigated by incorporating meta-learning into region-based detection frameworks. Despite its success, the said paradigm is constrained by several factors, such as \textit{(i)} low-quality region proposals for novel classes and \textit{(ii)} negligence of the inter-class correlation among different classes. Such limitations hinder the generalization of base-class knowledge for the detection of novel-class objects. In this work, we design Meta-DETR, a novel few-shot detection framework that incorporates correlational aggregation for meta-learning into DETR detection frameworks. Meta-DETR works entirely at image level without any region proposals, which circumvents the constraint of inaccurate proposals in prevalent few-shot detection frameworks. Besides, Meta-DETR can simultaneously attend to multiple support classes within a single feed-forward. This unique design allows capturing the inter-class correlation among different classes, which significantly reduces the misclassification of similar classes and enhances knowledge generalization to novel classes. Experiments over multiple few-shot object detection benchmarks show that the proposed Meta-DETR outperforms state-of-the-art methods by large margins. The implementation codes will be released.

\end{abstract}

\begin{figure}[t!] 
\begin{center}
   \includegraphics[width=1.0\linewidth]{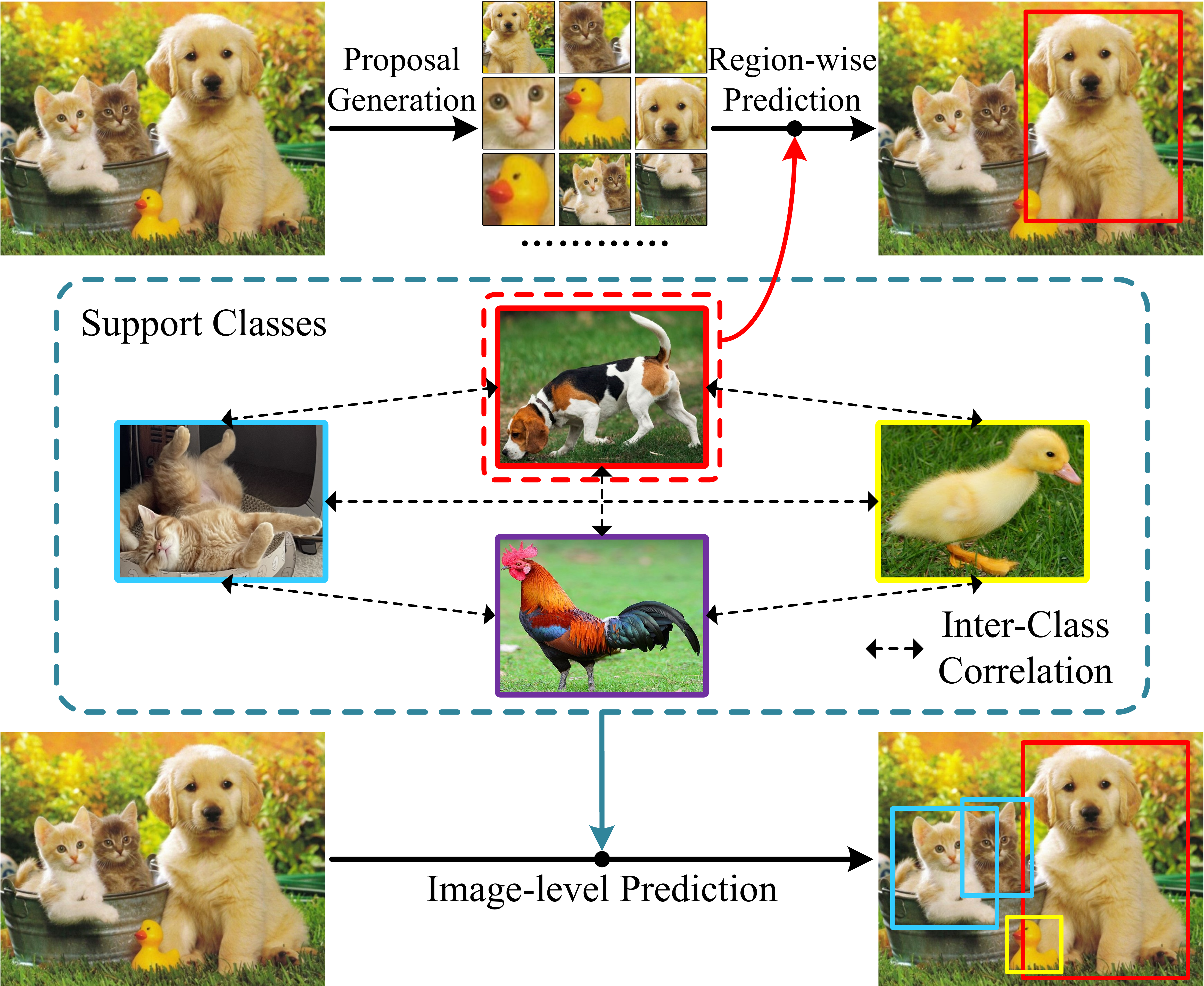}
\end{center}
\vspace*{-2.66mm}
\caption{
Comparison of few-shot object detection pipelines: Prior works (upper part) perform region-level detection, which are often constrained by inaccurate region proposals for novel classes. Besides, they can only deal with one support class at one go and overlook the correlation among different classes. The proposed Meta-DETR (lower part) works at image level without any proposals. It captures inter-class correlation by learning from multiple support classes simultaneously, which suppresses confusion among similar classes and enhances model generalization greatly.
}
\label{fig:fig1}
\vspace*{-1.5mm}
\end{figure}

\begin{figure*}[t!] 
\vspace*{-1.66mm}
\begin{center}
   \includegraphics[width=1.0\linewidth]{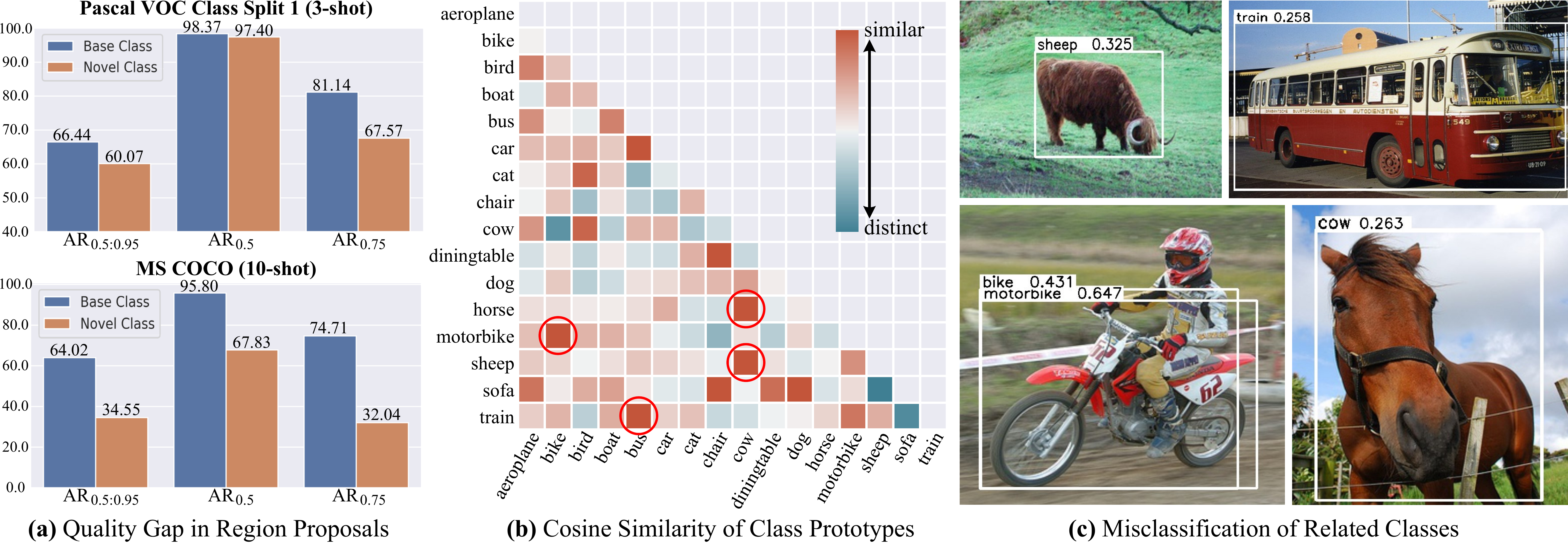}
\end{center}
\vspace*{-3.666mm}
\caption{Existing few-shot detection frameworks tend to suffer from inaccurate region proposals and under-exploitation of inter-class correlation. Due to very limited training samples, the proposal quality (measured by Average Recall on top 1000 proposals) for novel classes is clearly lower than that of base classes, as shown in (a). This hinders the knowledge generalization to novel classes. Additionally, object classes with similar appearances are highly correlated in feature space as shown in (b), which tend to be misclassified if the learning does not incorporate the correlation among them, as illustrated in (c).}
\label{fig:fig2_limitation}
\vspace*{-0.5mm}
\end{figure*}

\blfootnote{$\dagger$\, denotes equal contribution.}
\blfootnote{$*$\, denotes corresponding author.}

\section{Introduction}

Computer vision has experienced significant progress in recent years. However, there still exists a huge gap between current computer vision techniques and the human visual system in learning new concepts from very few examples: most existing methods require a large amount of annotated samples, while humans can effortlessly recognize a new concept even with very few instructions~\citep{Landau1988TheIO}. Such human-like capability to generalize from limited examples is highly desirable for machine vision systems, especially when sufficient training samples are unavailable or their annotations are hard to obtain.

In this work, we explore the challenging task of \textit{few-shot object detection}, which requires detecting novel objects with only a few training samples. With extremely limited supervision from annotated samples, the key is to exploit knowledge from base classes and generalize it to novel classes. To this end, many works~\citep{FewshotReweighting,metarcnn,FSDetView,fsod,DenseRelationDistillation} incorporate meta-learning into generic object detection frameworks, mostly Faster R-CNN~\citep{FasterRCNN}, and have achieved very promising results.

Despite their success, there still exist two underlying limitations that hinder better exploitation of base-class knowledge, as illustrated in Fig.\,\ref{fig:fig2_limitation}.
\textit{First}, region-based detection frameworks rely on region proposals to produce final predictions, thus are sensitive to low-quality region proposals. Unfortunately, as investigated by \citet{fsod} and \citet{CoRPN}, it is not easy to produce high-quality region proposals for novel classes with limited supervision under the few-shot detection setup. Such a gap in the quality of region proposals obstructs the generalization from base classes to novel classes.
\textit{Second}, most existing meta-learning-based approaches~\citep{FewshotReweighting,metarcnn,fsod,FSDetView} adopt `feature reweighting' or its variants to aggregate query and support features, which can only deal with one support class (\textit{i.e.}, target class to detect) at a time and essentially treat each support class independently.
Without seeing multiple classes within a single feed-forward, they largely overlook the important inter-class correlation among different support classes. This limits the ability to distinguish similar classes (\textit{e.g.}, distinguishing from cows and sheep) and to generalize from related classes (\textit{e.g.}, learning to detect cows by generalizing from detecting sheep).

To mitigate the above limitations, we design Meta-DETR, an innovative few-shot object detector that achieves meta-learning at image level and at the same time explicitly exploits the inter-class correlation among different support classes. To our best knowledge, this is the first work that explores incorporating meta-learning into the recently proposed DETR detection frameworks~\citep{DETR,DeformableDETR}, which can skip proposal generation and directly perform detection at image level. With image-level meta-learning, the proposed Meta-DETR lifts the constraint of inaccurate region proposals as in prevalent few-shot detection frameworks. In addition, as shown in Fig.\,\ref{fig:fig1}, Meta-DETR can attend to multiple support classes at one go instead of class-by-class meta-learning with repeated runs as in most existing methods.
By integrating detection tasks that involve multiple classes into meta-learning, Meta-DETR can explicitly leverage the inter-class correlation, including \textit{(i)} the inter-class commonality to facilitate generalization among related classes and \textit{(ii)} the inter-class uniqueness to reduce misclassification among similar classes.

In summary, the contributions of this work are threefold.
\textit{First}, we propose Meta-DETR, an innovative few-shot object detection framework that incorporates meta-learning into DETR detection frameworks. Being the first pure image-level meta-detector, Meta-DETR circumvents the gap of inaccurate region proposals for novel-class objects, enabling better generalization to novel classes.
\textit{Second}, we design a novel correlational aggregation module for few-shot object detection, which allows aggregating query features with multiple support classes simultaneously. It enables effective exploitation of the inter-class correlation, which greatly reduces misclassification and enhances model generalization.
\textit{Third}, extensive experiments show that, without bells and whistles, the proposed Meta-DETR outperforms state-of-the-art methods by large margins.

\begin{figure*}[t!] 
\begin{center}
   \includegraphics[width=1.0\linewidth]{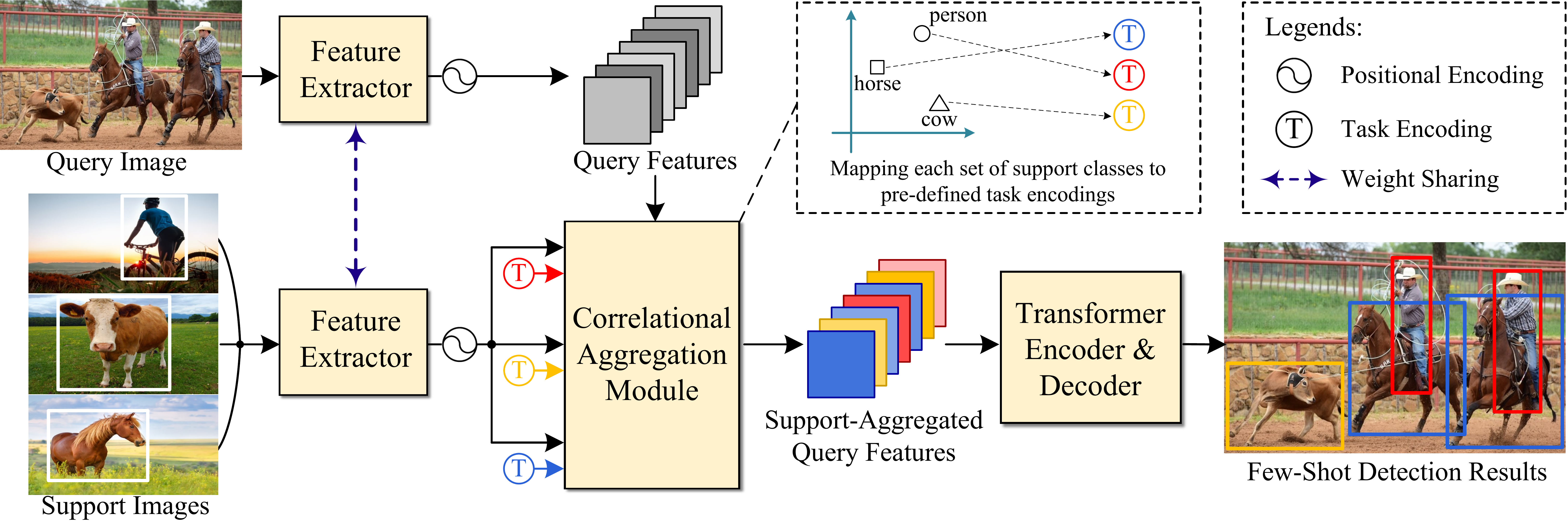}
\end{center}
\vspace*{-4.0mm}
   \caption{
   The framework of the proposed Meta-DETR. Query and support images are processed by a weight-shared feature extractor to produce query and support features. To leverage the inter-class correlation in meta-learning, the correlational aggregation module (CAM) first matches the query features with multiple support classes simultaneously, then introduces task encodings to differentiate these support classes. Finally, few-shot detection results are obtained via a class-agnostic transformer architecture that predicts objects' locations and corresponding task encodings.
   }
\label{fig:fig3_architecture}
\vspace*{-1.0mm}
\end{figure*}

\section{Related Work}

\noindent\textbf{Object Detection\;\;}
Generic object detection~\citep{Liu2019DeepLF} is a joint task on object localization and classification. Modern object detectors are mostly region-based and can be broadly classified into two categories: two-stage and single-stage detectors. Two-stage detectors include Faster R-CNN~\citep{FasterRCNN} and its variants~\citep{relationnetworks_detection,CascadeRCNN,CADNet}, which first adopt a Region Proposal Network (RPN) to generate region proposals, and then produce final predictions based on the proposals. Differently, single-stage detectors~\citep{SSD,YOLO9000,RefineDet} employ densely placed anchors as region proposals and directly make predictions over them.
Recently, another line of research featuring DETR~\citep{DETR} and its variants~\citep{DeformableDETR,up-detr} has received vast attention, thanks to the merits of pure image-level framework, fully end-to-end pipeline, and comparable or even better performance.
However, these aforementioned generic detectors still heavily rely on large amounts of annotated training samples, thus will suffer from drastic performance drop when directly applied to few-shot object detection.

\smallskip
\noindent\textbf{Few-Shot Object Detection\;\;}
Existing works on few-shot object detection can be categorized into two paradigms: transfer learning and meta-learning. Transfer-learning-based methods include LSTD\,\citep{LSTD}, TFA\,\citep{fsdet}, MPSR\,\citep{MPSR}, and FSCE\,\citep{fsce}, where novel concepts are learned via fine-tuning. Differently, meta-learning-based methods\;\citep{FewshotReweighting,metarcnn,metadet,incrementalfsdet,FSDetView,fsod,DenseRelationDistillation} extract knowledge that can generalize across various tasks via `learning to learn', \textit{i.e.}, learning a class-agnostic predictor on various auxiliary tasks.

Our proposed Meta-DETR falls under the umbrella of meta-learning, but differs from existing approaches by achieving image-level meta-learning and effectively leveraging the correlation among various support classes. To the best of our knowledge, Meta-DETR is the first work that incorporates meta-learning into the recently proposed DETR frameworks; It is also the pioneering work to explicitly integrate the inter-class correlation among support classes into meta-learning-based few-shot object detection frameworks.

\section{Preliminaries}

\noindent\textbf{Problem Definition\;\;}
Given two sets of classes $\mathcal{C}_{\rm base}$ and $\mathcal{C}_{\rm novel}$, where $\mathcal{C}_{\rm base} \cap \mathcal{C}_{\rm novel}=\varnothing$, a few-shot object detector aims at detecting objects of $\mathcal{C}_{\rm base} \cup \mathcal{C}_{\rm novel}$ by learning from a base dataset $\mathcal{D}_{\rm base}$ with abundant annotated objects of $\mathcal{C}_{\rm base}$ and a novel dataset $\mathcal{D}_{\rm novel}$ with very few annotated objects of $\mathcal{C}_{\rm novel}$. In the task of $K$-shot object detection, there are exactly $K$ annotated objects for each novel class in $\mathcal{D}_{\rm novel}$.

\vspace{+0.5mm}
\smallskip
\noindent\textbf{Rethink Region-Based Detection Frameworks\;\;}
Most existing works on few-shot object detection are developed on top of Faster R-CNN\,\cite{FasterRCNN}, a region-based object detector, thanks to its robust performance and easy optimization. However, by relying on region proposals to produce detection results, these approaches are inevitably constrained by the inaccurate proposals for novel classes due to very limited supervision under the few-shot detection setup. As illustrated in Fig.\;\ref{fig:fig2_limitation}(a), there is a clear gap in the quality of region proposals for base and novel classes, hindering region-based detection frameworks from fully exploiting base-class knowledge to generalize to novel classes. Though several studies\;\cite{fsod,CoRPN} attempt to acquire more accurate region proposals, this issue still remains as it is rooted in the region-based detection frameworks under the few-shot learning setup.

\smallskip
\vspace{+0.5mm}
\noindent\textbf{Rethink Meta-Learning via Feature Reweighting\;\;}
To meta-learn a class-agnostic detector that can generalize across various classes, most existing methods~\citep{FewshotReweighting,metarcnn,fsod,FSDetView} adopt `feature reweighting' or its variants to aggregate query features with support class information, acquiring class-specific meta-features to detect objects corresponding to the support class. However, such aggregation approaches can deal with only one support class within each feed-forward process, \textit{i.e.}, $C$ repeated runs are required to detect $C$ classes within each query image. More importantly, by treating each support class independently, `feature reweighting' overlooks the essential inter-class correlation among different support classes. As shown in Fig.\,\ref{fig:fig2_limitation}(b), many object classes with similar appearances are highly correlated. Intuitively, their correlation can effectively facilitate the distinction and the generalization among similar classes. However, as shown in Fig.\,\ref{fig:fig2_limitation}(c), in existing methods, we observe that objects misclassified as highly correlated classes constitute a major source of error due to the negligence of inter-class correlation.

\section{Meta-DETR}   \label{sec:meta-detr}

%This section provides a detailed description of the proposed Meta-DETR, including its network architecture, training objective, as well as the learning and inference procedure.

\subsection{Model Overview}

Fig.\,\ref{fig:fig3_architecture} presents the architecture of the proposed Meta-DETR. Motivated by previous discussions, Meta-DETR employs the recently proposed Deformable DETR\;\cite{DeformableDETR}, a fully end-to-end Transformer-based\;\cite{transformer} detector, as the basic detection framework to bypass the constraint of region-wise prediction. Besides, during meta-learning, Meta-DETR aggregates query features with multiple support classes simultaneously, thus can exploit the inter-class correlation among different classes to reduce misclassification and boost generalization.

Specifically, given a query image and a set of support images with instance annotations, a weight-shared feature extractor first encodes them into the same feature space. Subsequently, the correlational aggregation module (CAM), which will be introduced in detail later, performs matching between the query features and the set of support classes. CAM further maps the set of support classes to a set of pre-defined task encodings that differentiate these support classes in a class-agnostic manner. Finally, detection results are obtained via a transformer architecture that predicts objects' locations and corresponding task encodings. As the detection targets are dynamically determined by support classes and their mappings to task encodings, the proposed Meta-DETR is trained as a meta-learner to extract generalizable knowledge not specific to certain classes.

\subsection{Correlational Aggregation Module}

\begin{figure}[t!] 
\begin{center}
   \includegraphics[width=1.0\linewidth]{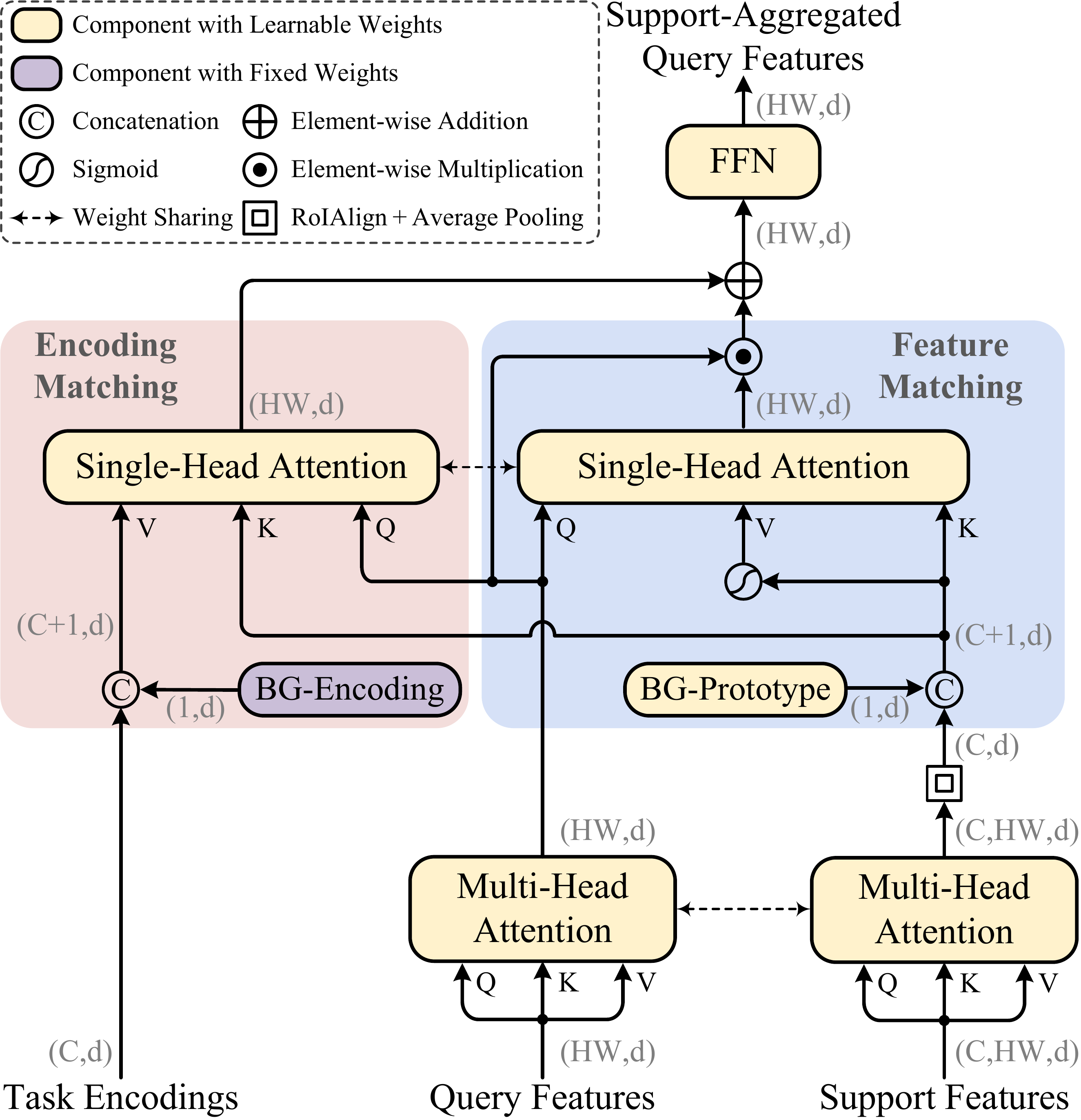}
\end{center}
\vspace*{-3.0mm}
   \caption{The architecture of the Correlational Aggregation Module (CAM). It performs two matching processes: feature matching filters out query features that are unrelated to support classes, while encoding matching maps support classes to a set of pre-defined task encodings that differentiate the support classes in a class-agnostic manner.
   }
\label{fig:fig4_CAM}
\vspace*{-0.0mm}
\end{figure}

The correlational aggregation module (CAM) is the key component in Meta-DETR, which aggregates query features with support classes for the subsequent class-agnostic prediction.
CAM differs from existing aggregation methods in that it can aggregate multiple support classes simultaneously, which enables it to capture their inter-class correlation to reduce misclassification and enhance model generalization. Specifically, as illustrated in Fig.\;\ref{fig:fig4_CAM}, given the query and support features, a weight-shared multi-head attention module first encodes them into the same feature space, and the prototype for each support class is obtained by applying RoIAlign\;\cite{MaskRCNN} followed by average pooling on the support features. CAM then performs feature matching and encoding matching, which will be elaborated in the remainder of this subsection, to match the query features with support features and task encodings, respectively. Their results are summed together and processed by a feed-forward network (FFN) to produce the final output.

\begin{table*}[t]
\vspace*{-2.50mm}
\begin{center}
\centering
\setlength{\tabcolsep}{4.98pt}
\resizebox{1.0\textwidth}{!}{
\begin{tabular}[t]{l|ccccc|ccccc|ccccc|c}
\toprule[1.1pt]
&\multicolumn{5}{c|}{Class Split 1} & \multicolumn{5}{c|}{Class Split 2} & \multicolumn{5}{c|}{Class Split 3} & \multirow{2}*{\textit{Avg.}} \\

 \cmidrule{1-16}

Method $\backslash$ Shot  & 1 & 2 & 3 & 5 & 10 & 1 & 2 & 3 & 5 & 10 & 1 & 2 & 3 & 5 & 10 & \\\midrule[0.68pt]

\textit{Results over a single run:} & & & & & & & & & & & & & & &  \\

LSTD\;\cite{LSTD} & 8.2 & 1.0 & 12.4 & 29.1 & 38.5 & 11.4 & 3.8 & 5.0 & 15.7 & 31.0 & 12.6 & 8.5 & 15.0 & 27.3 & 36.3 & 17.1 \\

RepMet\;\cite{RepMet}\,$\ddag$ & 26.1 & 32.9 & 34.4 & 38.6 & 41.3 & 17.2 & 22.1 & 23.4 & 28.3 & 35.8 & 27.5 & 31.1 & 31.5 & 34.4 & 37.2 & 30.8 \\

%NP-RepMet\;\cite{NP-RepMet} & 37.8 & 40.3 & 41.7 & 47.3 & 49.4 & 41.6 & 43.0 & 43.4 & 47.4 & 49.1 & 33.3 & 38.0 & 39.8 & 41.5 & 44.8 & ??? \\

Meta-YOLO\;\cite{FewshotReweighting} & 14.8 & 15.5 & 26.7 & 33.9 & 47.2 & 15.7 & 15.3 & 22.7 & 30.1 & 40.5 & 21.3 & 25.6 & 28.4 & 42.8 & 45.9 & 28.4 \\

Meta\,Det\;\cite{metadet} & 18.9 & 20.6 & 30.2 & 36.8 & 49.6 & 21.8 & 23.1 & 27.8 & 31.7 & 43.0 & 20.6 & 23.9 & 29.4 & 43.9 & 44.1 & 31.0 \\

Meta R-CNN\;\cite{metarcnn} & 19.9 & 25.5 & 35.0 & 45.7 & 51.5 & 10.4 & 19.4 & 29.6 & 34.8 & 45.4 & 14.3 & 18.2 & 27.5 & 41.2 & 48.1 & 31.1 \\

% TFA w/ fc\;\cite{fsdet}\,$\ddag$ & 36.8 & 29.1 & 43.6 & 55.7 & 57.0 & 18.2 & 29.0 & 33.4 & 35.5 & 39.0 & 27.7 & 33.6 & 42.5 & 48.7 & 50.2 \\

TFA w/ cos\;\cite{fsdet}\,$\ddag$ & 39.8 & 36.1 & 44.7 & 55.7 & 56.0 & 23.5 & 26.9 & 34.1 & 35.1 & 39.1 & 30.8 & 34.8 & 42.8 & 49.5 & 49.8 & 39.9 \\

MPSR\;\cite{MPSR}\,$\ddag$ & 41.7 & 43.1 & 51.4 & 55.2 & 61.8 & 24.4 & 29.5 & 39.2 & 39.9 & 47.8 & 35.6 & 40.6 & 42.3 & 48.0 & 49.7 & 43.3 \\

TFA w/ cos + Halluc\;\cite{Halluc_FSD}\,$\ddag$ & 45.1 & 44.0 & 44.7 & 55.0 & 55.9 & 23.2 & 27.5 & 35.1 & 34.9 & 39.0 & 30.5 & 35.1 & 41.4 & 49.0 & 49.3 & 40.6 \\

CME\;\cite{CME}\,$\ddag$ & 41.5 & 47.5 & 50.4 & 58.2 & 60.9 & 27.2 & 30.2 & 41.4 & 42.5 & 46.8 & 34.3 & 39.6 & 45.1 & 48.3 & 51.5 & 44.4 \\

SRR-FSD\;\cite{SRR-FSD}\,$\ddag$\,$\uplus$ & \textbf{47.8} & 50.5 & 51.3 & 55.2 & 56.8 & 32.5 & 35.3 & 39.1 & 40.8 & 43.8 & 40.1 & 41.5 & 44.3 & 46.9 & 46.4 & 44.8\\

FSCE\;\cite{fsce}\,$\ddag$ & 44.2 & 43.8 & 51.4 & \textbf{61.9} & 63.4 & 27.3 & 29.5 & 43.5 & 44.2 & 50.2 & 37.2 & 41.9 & 47.5 & 54.6 & 58.5 & 46.6\\

\rowcolor{black!6} Meta-DETR (Ours) & 40.6 & \textbf{51.4} & \textbf{58.0} & 59.2 & \textbf{63.6} & \textbf{37.0} & \textbf{36.6} & \textbf{43.7} & \textbf{49.1} & \textbf{54.6} & \textbf{41.6} & \textbf{45.9} & \textbf{52.7} & \textbf{58.9} & \textbf{60.6} & \textbf{50.2} \\\midrule[0.68pt]

\textit{Results averaged over multiple random runs:} & & & & & & & & & & & & & & & \\

FRCN-ft-full\;\cite{FasterRCNN}\,$\ddag$ & 9.9 & 15.6 & 21.6 & 28.0 & 35.6 & 9.4 & 13.8 & 17.4 & 21.9 & 29.8 & 8.1 & 13.9 & 19.0 & 23.9 & 31.0 & 19.9 \\

Deformable-DETR-ft-full\;\cite{DeformableDETR}\,$\ddag$ & 5.6 & 13.3 & 21.7 & 34.2 & 45.0 & 10.9 & 13.0 & 18.4 & 27.3 & 39.4 & 7.3 & 16.6 & 20.8 & 32.2 & 41.8 & 23.2 \\

% TFA w/ fc\;\cite{fsdet}\,$\ddag$ & 22.9 & 34.5 & 40.4 & 46.7 & 52.0 & 16.9 & 26.4 & 30.5 & 34.6 & 39.7 & 15.7 & 27.2 & 34.7 & 40.8 & 44.6 \\

TFA w/ cos\;\cite{fsdet}\,$\ddag$ & 25.3 & 36.4 & 42.1 & 47.9 & 52.8 & 18.3 & 27.5 & 30.9 & 34.1 & 39.5 & 17.9 & 27.2 & 34.3 & 40.8 & 45.6 & 34.7 \\

FsDetView\;\cite{FSDetView} & 24.2 & 35.3 & 42.2 & 49.1 & 57.4 & 21.6 & 24.6 & 31.9 & 37.0 & 45.7 & 21.2 & 30.0 & 37.2 & 43.8 & 49.6 & 36.7 \\

MPSR\;\cite{MPSR}\,$\ddag\,{\triangle}$ & 34.7 & 42.6 & 46.1 & 49.4 & 56.7 & 22.6 & 30.5 & 31.0 & 36.7 & 43.3 & 27.5 & 32.5 & 38.2 & 44.6 & 50.0 & 39.1 \\

DCNet\;\cite{DenseRelationDistillation}\,$\ddag$ & 33.9 & 37.4 & 43.7 & 51.1 & 59.6 & 23.2 & 24.8 & 30.6 & 36.7 & 46.6 & 32.3 & 34.9 & 39.7 & 42.6 & 50.7 & 39.2 \\

FSCE\;\cite{fsce}\,$\ddag$ & 32.9 & 44.0 & 46.8 & 52.9 & 59.7 & 23.7 & 30.6 & \textbf{38.4} & 43.0 & 48.5 & 22.6 & 33.4 & 39.5 & 47.3 & 54.0 & 41.2 \\

\rowcolor{black!6} Meta-DETR (Ours) & \textbf{35.1} & \textbf{49.0} & \textbf{53.2} & \textbf{57.4} & \textbf{62.0} & \textbf{27.9} & \textbf{32.3} & \textbf{38.4} & \textbf{43.2} & \textbf{51.8} & \textbf{34.9} & \textbf{41.8} & \textbf{47.1} & \textbf{54.1} & \textbf{58.2} & \textbf{45.8} \\

\bottomrule[1.1pt]
\end{tabular}}
\end{center}
\vspace*{-3.99mm}
\caption{Few-shot detection performance (mAP@0.5) on Pascal VOC \textit{test\,07} for novel classes. $\ddag$ indicates methods using multi-scale features. $\triangle$ indicates re-evaluated results using official codes. $\uplus$ indicates usage of external data.}
\label{tab:Performance_VOC_novel}
\vspace*{-1.5mm}
\end{table*}

\smallskip
\vspace{+1.00mm}
\noindent\textbf{Feature Matching \;}
Feature matching is accomplished by a single-head attention mechanism. Specifically, given a query feature map $\mathbf{Q} \in \mathbb{R}^{HW \times d}$ and the support class prototypes $\mathbf{S} \in \mathbb{R}^{C \times d}$, the matching coefficients are obtained via:
\begin{equation}
\mathbf{A} = {\rm Attn}(\mathbf{Q}, \mathbf{S}) = {\rm Softmax}(\frac{(\mathbf{Q W})(\mathbf{S W})^{\rm T}}{\sqrt{d}}),
\end{equation}
where $HW$ is the spatial size, $C$ is the number of support classes, $d$ is the feature dimensionality, and $\mathbf{W}$ is a linear projection shared by $\mathbf{Q}$ and $\mathbf{S}$, which ensures they are embedded into the same feature space. Subsequently, the output of the feature matching module can be obtained via:
\begin{equation}
\mathbf{Q}_{\mathbf{F}} = \mathbf{A} \sigma(\mathbf{S}) \odot \mathbf{Q},
\end{equation}
where $\sigma(\cdot)$ denotes sigmoid function and $\odot$ denotes Hadamard product. $\sigma(\mathbf{S})$ serves as feature filters for each individual support class with the function of extracting only class-related features from query features. By applying the matching coefficients $\mathbf{A}$ to $\sigma(\mathbf{S})$, we filter out features not matched to any support class, producing a feature map $\mathbf{Q_F}$ that highlights objects belonging to the given support classes.

\smallskip
\vspace{+1.00mm}
\noindent\textbf{Encoding Matching\;\;}
To achieve meta-learning that requires class-agnostic prediction, we introduce a set of pre-defined task encodings and map the given support classes to these task encodings, so that final predictions can be made on the task encodings instead of specific classes. We implement task encodings $\mathbf{T} \in \mathbb{R}^{C \times d}$ with sinusoidal functions, following the positional encodings of the Transformer\,\cite{transformer}. Encoding matching uses the same matching coefficients as feature matching, and the matched encodings $\mathbf{Q_E}$ are obtained via:
\begin{equation}
\mathbf{Q}_{\mathbf{E}} = \mathbf{A} \mathbf{T}.
\end{equation}

\smallskip
\noindent\textbf{Modeling Background for Open-Set Prediction\;\;}
Object detection features an open-set setup where background, which does not belong to any of the target classes, often takes up most of the space in a query image. Therefore, as shown in Fig.\,\ref{fig:fig4_CAM}, we additionally introduce a learnable prototype and a corresponding task encoding (fixed to zeros), denoted as BG-Prototype and BG-Encoding respectively, to explicitly model the background class. This eliminates the matching ambiguity when query does not match any of the given support classes.

\subsection{Training Objective} \label{sec:TrainingObjective}

\noindent\textbf{Target Generation\;\;}
The detection targets of Meta-DETR are dynamically determined by the support classes and their mappings to task encodings. Concretely, given a query image, $C$ support images representing different support classes are randomly sampled. Only ground truth objects belonging to the sampled support classes are kept as detection targets. Besides, the classification target for each object is the task encoding of the ground truth class instead of the ground truth class itself. We empirically set $C$ as 5 according to our ablation study on this hyper-parameter in Fig.\;\ref{fig:fig5_num_supp_class}.

\smallskip
\noindent\textbf{Loss Function\;\;}
The loss functions for our proposed Meta-DETR follow Deformable DETR\;\cite{DeformableDETR}, which adopts a set-based Hungarian loss that forces unique predictions for each object via bipartite matching. Following Meta R-CNN\;\cite{metarcnn}, we additionally introduce a cosine similarity cross-entropy loss\;\cite{CloserFewshotClassification} to classify the class prototypes obtained by our designed CAM. It encourages prototypes of different classes to be distinguished from each other. Please refer to appendix for a detailed description of the loss functions.

\subsection{Training and Inference Procedure}
\noindent\textbf{Two-Stage Training Procedure\;\;}
The training procedure consists of two stages.
The first stage is \textit{base training stage}. During this stage, the model is trained on the base dataset $\mathcal{D}_{\rm base}$ with abundant training samples for each base class.
The second stage is \textit{few-shot fine-tuning stage}. In this stage, we train the model on both base and novel classes with limited training samples. Only $K$ object instances are available for each novel category in $K$-shot object detection. Following prior works~\cite{metarcnn,fsdet,FSDetView}, we also include objects from base classes to prevent performance drop for base classes. In both stages, the network is optimized in an end-to-end manner with the same training objective described in Section\;\ref{sec:TrainingObjective}.

\smallskip
\noindent\textbf{Efficient Inference\;\;}
Unlike the training stage, there is no need to repeatedly sample support images and extract their features. We can first compute the prototypes for each support class once and for all, then directly use them for every query image to predict. This promises the efficient inference of our proposed Meta-DETR.

\vspace{-1.50mm}
\section{Experiments}

\subsection{Datasets}

We follow the well-established data setups for few-shot object detection~\cite{FewshotReweighting,fsdet}. Concretely, two widely used few-shot object detection benchmarks are adopted in our experiments.

\smallskip
\noindent
\textbf{Pascal VOC}~\cite{PascalVOC}
consists of images with object annotations of 20 classes. We use \textit{trainval\,07+12} for training and perform evaluations on \textit{test\,07}. We use 3 novel\,/\,base class splits, \textit{i.e.}, (“bird”, “bus”, “cow”, “motorbike”, “sofa”\;/\;others), (“aeroplane”, “bottle”,“cow”,“horse”,“sofa”\;/\;others) and (“boat”, “cat”, “motorbike”,“sheep”, “sofa”\;/\;others). The number of shots is set to 1, 2, 3, 5 and 10.  Mean average precision (mAP) at IoU threshold 0.5 is used as the evaluation metric. Results are averaged over 10 randomly sampled support datasets.

\smallskip
\vspace{+0.15mm}
\noindent
\textbf{MS COCO}~\cite{MSCOCO}
is a more challenging object detection dataset, which contains 80 classes including those 20 classes in Pascal VOC. We adopt the 20 shared classes as novel classes, and adopt the remaining 60 classes as base classes. The number of shots is set to 1, 3, 5, 10, and 30. We use \textit{train\,2017} for training, and perform evaluations on \textit{val\,2017}. Standard evaluation metrics for MS COCO are adopted. Results are averaged over 5 randomly sampled support datasets.

\subsection{Implementation Details} \label{sec:imple_details}

We adopt the commonly used ResNet-101\,\cite{resnet} as the feature extractor. The network architectures and hyper-parameters remain the same as Deformable DETR\;\cite{DeformableDETR}. We implement our model in single-scale version for fair comparison with other works. We also follow FsDetView\;\cite{FSDetView} to implement the aggregation with a slightly more complex scheme compared with solely feature reweighting.
Following Deformable DETR, we train our model with 8\,x Nvidia V100 GPUs, using the AdamW\,\cite{AdamW} optimizer with an initial learning rate of $\rm 2\!\times\!10^{-4}$ and a weight decay of $\rm 1\!\times\!10^{-4}$. Batch size is set to 32. In the base training stage, we train the model for 50 epochs for both Pascal VOC and MS COCO. Learning rate is decayed at the 45$^{\rm th}$ epoch by 0.1. In the few-shot fine-tuning stage, the same settings are applied to fine-tune the model until convergence.

\subsection{Comparison with State-of-the-Art Methods}

\noindent\textbf{Pascal VOC\;\;}
Table\;\ref{tab:Performance_VOC_novel} shows the few-shot detection performance for novel classes of Pascal VOC. It can be seen that Meta-DETR consistently outperforms existing methods across various setups. With multiple runs over randomly sampled support datasets to reduce randomness, our method achieves the best average performance across all setups, with a large margin of +\,4.6\% mAP compared with the second-best. The strong performance demonstrates the superiority and robustness of our proposed method.

\begin{table}[t!]
\vspace*{-2.5mm}
\begin{center}
\centering
\setlength{\tabcolsep}{2.486pt}
\resizebox{0.474\textwidth}{!}{
\begin{tabular}[t]{ c | l | ccc }
\toprule[1.0pt]

Shot & Method & AP & AP$_{\rm 0.5}$ & AP$_{\rm 0.75}$  \\

\midrule[0.68pt]

\multirow{5}{*}{1} & FRCN-ft-full\;\cite{FasterRCNN}\,$\ddag$\,$\S$ & 1.7 & 3.3 & 1.6 \\

& Deformable-DETR-ft-full\,\cite{DeformableDETR}\,$\S$ & 1.8 & 3.1 & 1.8 \\

& TFA w/ cos\;\cite{fsdet}\,$\ddag$\,$\S$ & 1.9 & 3.8 & 1.7 \\

& TFA\,w/\,cos\,+\,Halluc\,\cite{Halluc_FSD}\,$\ddag$ & 3.8 & 6.5 & 4.3 \\

& \cellcolor{black!6}Meta-DETR (Ours)\,$\S$ & \cellcolor{black!6}\textbf{7.5} & \cellcolor{black!6}\textbf{12.5} & \cellcolor{black!6}\textbf{7.7} \\

\midrule[0.68pt]

\multirow{5}{*}{3} & FRCN-ft-full\;\cite{FasterRCNN}\,$\ddag$\,$\S$ & 3.7 & 7.1 & 3.5 \\

& Deformable-DETR-ft-full\,\cite{DeformableDETR}\,$\S$ & 4.9 & 7.8 & 5.1 \\

& TFA w/ cos\;\cite{fsdet}\,$\ddag$\,$\S$ & 5.1 & 9.9 & 4.8 \\

& TFA\,w/\,cos\,+\,Halluc\,\cite{Halluc_FSD}\,$\ddag$ & 6.9 & 12.6 & 7.0 \\

& \cellcolor{black!6}Meta-DETR (Ours)\,$\S$ & \cellcolor{black!6}\textbf{13.5} & \cellcolor{black!6}\textbf{21.7} & \cellcolor{black!6}\textbf{14.0} \\

\midrule[0.68pt]

\multirow{5}{*}{5} & FRCN-ft-full\;\cite{FasterRCNN}\,$\ddag$\,$\S$ & 4.6 & 8.7 & 4.4 \\

& Deformable-DETR-ft-full\,\cite{DeformableDETR}\,$\S$ & 7.4 & 12.3 & 7.7 \\

& TFA w/ cos\;\cite{fsdet}\,$\ddag$\,$\S$ & 7.0 & 13.3 & 6.5 \\

& FsDetView\;\cite{FSDetView}\,$\S$ & 10.7 & 24.5 & 6.7 \\

& \cellcolor{black!6}Meta-DETR (Ours)\,$\S$ & \cellcolor{black!6}\textbf{15.4} & \cellcolor{black!6}\textbf{25.0} & \cellcolor{black!6}\textbf{15.8} \\

\midrule[0.68pt]

\multirow{14}{*}{10} & FRCN-ft-full\;\cite{FasterRCNN}\,$\ddag$\,$\S$ & 5.5 & 10.0 & 5.5 \\

& Deformable-DETR-ft-full\,\cite{DeformableDETR}\,$\S$ & 11.7 & 19.6 & 12.1 \\

& Meta-YOLO\;\cite{FewshotReweighting} & 5.6 & 12.3 & 4.6 \\

& Meta\,Det\;\cite{metadet} & 7.1 & 14.6 & 6.1 \\

& Meta R-CNN\;\cite{metarcnn} & 8.7 & 19.1 & 6.6 \\

& TFA w/ cos\;\cite{fsdet}\,$\ddag$\,$\S$ & 9.1 & 17.1 & 8.8 \\

& FSOD\;\cite{fsod} & 12.0 & 22.4 & 11.8 \\

& FsDetView\;\cite{FSDetView}\,$\S$ & 12.5 & 27.3 & 9.8 \\

& MPSR\;\cite{MPSR}\,$\ddag$ & 9.8 & 17.9 & 9.7  \\

& SRR-FSD\;\cite{SRR-FSD}\,$\ddag$ & 11.3 & 23.0 & 9.8 \\

& CME\;\cite{CME}\,$\ddag$ & 15.1 & 24.6 & 16.4 \\

& DCNet\;\cite{DenseRelationDistillation}\,$\ddag$\,$\S$ & 12.8 & 23.4 & 11.2 \\

& FSCE\;\cite{fsce}\,$\ddag$\,$\S$ & 11.1 & - & 9.8 \\

& \cellcolor{black!6}Meta-DETR (Ours)\,$\S$ & \cellcolor{black!6}\textbf{19.0} & \cellcolor{black!6}\textbf{30.5} & \cellcolor{black!6}\textbf{19.7}\\

\midrule[0.66pt]

\multirow{13}{*}{30} & FRCN-ft-full\;\cite{FasterRCNN}\,$\ddag$\,$\S$ & 7.4 & 13.1 & 7.4 \\

& Deformable-DETR-ft-full\,\cite{DeformableDETR}\,$\S$ & 16.3 & 27.2 & 16.7 \\

& Meta-YOLO\;\cite{FewshotReweighting} & 9.1 & 19.0 & 7.6\\

& Meta\,Det\;\cite{metadet} & 11.3 & 21.7 & 8.1\\

& Meta R-CNN\;\cite{metarcnn} & 12.4 & 25.3 & 10.8 \\

& TFA w/ cos\;\cite{fsdet}\,$\ddag$\,$\S$ & 12.1 & 22.0 & 12.0 \\

& FsDetView\;\cite{FSDetView}\,$\S$ & 14.7 & 30.6 & 12.2 \\

& MPSR\;\cite{MPSR}\,$\ddag$\ & 14.1 & 25.4 & 14.2 \\

& SRR-FSD\;\cite{SRR-FSD}\,$\ddag$ & 14.7 & 29.2 & 13.5 \\

& CME\;\cite{CME}\,$\ddag$ & 16.9 & 28.0 & 17.8 \\

& DCNet\;\cite{DenseRelationDistillation}\,$\ddag$\,$\S$ & 18.6 & 32.6 & 17.5 \\

& FSCE\;\cite{fsce}\,$\ddag$\,$\S$ & 15.3 & - & 14.2 \\

& \cellcolor{black!6}Meta-DETR (Ours)\,$\S$ & \cellcolor{black!6}\textbf{22.2} & \cellcolor{black!6}\textbf{35.0} & \cellcolor{black!6}\textbf{22.8} \\

\bottomrule[1.0pt]

\end{tabular}
}
\end{center}
\vspace*{-3.0mm}
\caption{Few-shot detection performance on MS COCO \textit{val\,2017} for novel classes.  $\ddag$ indicates methods using multi-scale features. $\S$ indicates results averaged on multiple runs.}
\label{tab:Performance_COCO_novel}
\vspace*{-2.0mm}
\end{table}

\vspace{+0.5mm}
\smallskip
\noindent\textbf{MS COCO \;\;}
Table\;\ref{tab:Performance_COCO_novel} shows the results on MS COCO. It can be seen that, although MS COCO is much more challenging than Pascal VOC with higher complexity like occlusions and large scale variations, Meta-DETR still outperforms all existing methods under all setups by even larger margins. This can be potentially attributed to the effective exploitation of the correlations among more classes in MS COCO. In addition, Meta-DETR performs exceptionally well compared with other region-based methods under the stricter metric AP$_{\rm 0.75}$, which implies our method can effectively lift the constraint of inaccurate region proposals, thus producing more accurate detection results.

\subsection{Ablation Studies}

We conduct comprehensive ablation studies to verify the effectiveness of our design choices. All results are averaged over 10 runs with different randomly sampled support datasets on the first class split of Pascal VOC.

\vspace{+0.5mm}
\smallskip
\noindent\textbf{Region-Level \textit{vs.} Image-Level \;\;}
From Table\;\ref{tab:Performance_VOC_novel} and Table\;\ref{tab:Performance_COCO_novel}, we can find that fine-tuning Deformable DETR (Deformable-DETR-ft-full) generally outperforms fine-tuning Faster R-CNN (FRCN-ft-full), especially in the MS COCO dataset, where it is much harder to obtain accurate region proposals for novel classes due to higher complexity (see Fig.\;\ref{fig:fig2_limitation}(a)). This observation aligns well with our insight that region-based frameworks tend to suffer from inaccurate regional proposals for novel classes. To further verify the superiority of image-level few-shot object detection, we adopt FsDetView\,\cite{FSDetView}, a state-of-the-art meta-learning-based few-shot detector built on top of Faster R-CNN, as a solid baseline to compare with our method. For a fair comparison, we add a deformable transformer to FsDetView to rule out the performance difference brought by the transformer architecture. Furthermore, we replace our proposed CAM in Meta-DETR with the feature aggregation module in FsDetView (denoted as Meta-DETR w/o CAM). As shown in Table\,\ref{tab:ablation1}, even with aligned network architecture and aggregation scheme, Meta-DETR w/o CAM still outperforms FsDetView\;+\;Deform\,Transformer under most setups. The results validate the superiority of solving few-shot object detection at image level.

\begin{figure*}[t!] 
\begin{center}
   \includegraphics[width=1.0\linewidth]{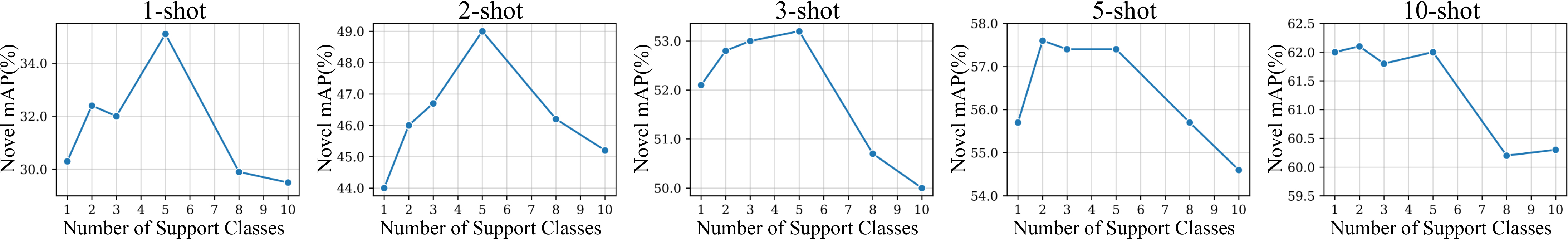}
\end{center}
\vspace*{-3.66mm}
   \caption{
   Ablation study over the number of support classes for correlational aggregation under different few-shot setups.
   %Results are averaged over 10 repeated runs on the first class split of Pascal VOC.
   }
\label{fig:fig5_num_supp_class}
\vspace*{-2.486mm}
\end{figure*}

\begin{table}[t]
\vspace*{-2.66mm}
\begin{center}
\centering
\setlength{\tabcolsep}{4.86pt}
\resizebox{0.474\textwidth}{!}{
\begin{tabular}[t]{l|ccccc}
\toprule[1.0pt]

& \multicolumn{5}{c}{Novel mAP@0.5} \\

Method $\backslash$ Shot & 1 & 2 & 3 & 5 & 10 \\

\midrule[0.56pt]

FsDetView\;\cite{FSDetView} & 24.2 & 35.3 & 42.2 & 49.1 & 57.4 \\

FsDetView\;+\;Deform\,Transformer & \textbf{28.0} & 36.3 & 41.8 & 48.9 & 57.4 \\

Meta-DETR w/o CAM & 27.2 & \textbf{42.1} & \textbf{50.5} & \textbf{52.9} & \textbf{59.3} \\

\bottomrule[1.0pt]

\end{tabular}
}
\end{center}
\vspace*{-3.66mm}
\caption{Performance comparison between region-level and image-level meta-learning-based few-shot object detection.}
\label{tab:ablation1}
\vspace*{-3.0mm}
\end{table}

\begin{table}[t]
\begin{center}
\centering
\setlength{\tabcolsep}{3.88pt}
\resizebox{0.474\textwidth}{!}{
\begin{tabular}[t]{l|ccc|ccccc}
\toprule[1.0pt]

\multirow{2}{*}{Method} & \multirow{2}{*}{CAM} & Modeling & \multirow{2}{*}{$C$} & \multicolumn{5}{c}{Novel mAP@0.5} \\

 & & Background &  & 1 & 2 & 3 & 5 & 10 \\

\midrule[0.56pt]

\multirow{4}{*}{Meta-DETR} &  &  & 1 & 27.2 & 42.1 & 50.5 & 52.9 & 59.3 \\

 & $\checkmark$ & $\checkmark$ & 1 & 30.3 & 44.0 & 52.1 & 55.7 & \textbf{62.0} \\

 & $\checkmark$ &  & 5 & 32.6 & 45.6 & 51.3 & 56.1 & 60.9 \\

 & \cellcolor{black!6}$\checkmark$ & \cellcolor{black!6}$\checkmark$ & \cellcolor{black!6}5 & \cellcolor{black!6}\textbf{35.1} & \cellcolor{black!6}\textbf{49.0} & \cellcolor{black!6}\textbf{53.2} & \cellcolor{black!6}\textbf{57.4} & \cellcolor{black!6}\textbf{62.0} \\

\bottomrule[1.0pt]

\end{tabular}
}
\end{center}
\vspace*{-3.66mm}
\caption{Ablation study to evaluate the effectiveness of our designed CAM and its design choices. $C$ denotes the number of support classes to aggregate simultaneously.}
\label{tab:ablation2}
\vspace*{-3.0mm}
\end{table}

\begin{table}[!t]
\begin{center}
\centering
\setlength{\tabcolsep}{3.486pt}
\resizebox{0.474\textwidth}{!}{
\begin{tabular}[t]{l|c|ccccc}
\toprule[1.0pt]

& & \multicolumn{5}{c}{Novel mAP@0.5} \\

Method $\backslash$ Shot & $C$ & 1 & 2 & 3 & 5 & 10 \\

\midrule[0.56pt]

FsDetView\;\cite{FSDetView} & 1 & 24.2 & 35.3 & 42.2 & 49.1 & 57.4 \\

FsDetView\;w/\;CAM & 5 & \textbf{30.1} & \textbf{41.1} & \textbf{45.2} & \textbf{51.4} & \textbf{57.5} \\

\bottomrule[1.0pt]

\end{tabular}
}
\end{center}
\vspace*{-3.66mm}
\caption{Ablation study on the effectiveness of our designed CAM on region-based detection frameworks. $C$ denotes the number of support classes to aggregate simultaneously.}
\label{tab:ablation3}
\vspace*{-1.68mm}
\end{table}

\begin{figure}[t] 
\begin{center}
   \includegraphics[width=1.0\linewidth]{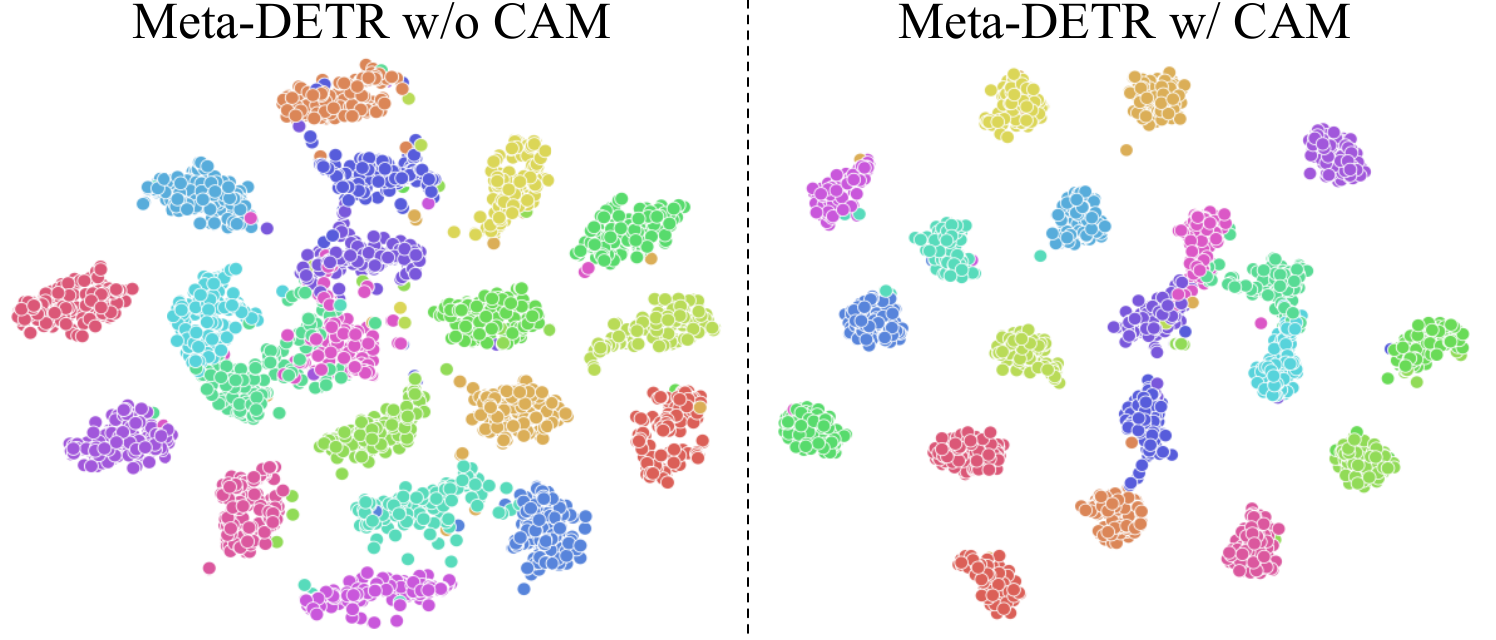}
\end{center}
\vspace*{-3.88mm}
   \caption{
   t-SNE visualization of objects learned in the feature space with and without our designed CAM. Results are obtained on split 1 of Pascal VOC under the 2-shot setup.
   }
\label{fig:fig6_tsne}
\vspace*{-1.88mm}
\end{figure}

\vspace{+0.5mm}
\smallskip
\noindent\textbf{Impact of Correlational Aggregation Module (CAM)\;\;}
As shown in Table\;\ref{tab:ablation2}, when incorporating CAM into our model, even if we keep the number of support classes as 1, which means CAM cannot explicitly leverage inter-class correlation among different support classes, CAM can still boost few-shot detection performance under all settings. This demonstrates CAM's strong capacity in aggregating query and support information. When multiple support classes are available, CAM can further enable the exploitation of their inter-class correlation to boost few-shot detection performance under lower-shot ($\leq$5) settings, especially under 1-shot (+\,4.8\%\;mAP) and 2-shot (+\,5.0\%\;mAP). No clear performance gain is observed for 10-shot, which implies that, when more training samples are available, the detector can already recognize novel classes and differentiate them from similar classes without explicitly modeling their inter-class correlation.
We also apply our designed CAM to the commonly used region-based meta-detector FsDetView and report the results in Table\;\ref{tab:ablation3}. Its steady performance gain demonstrates CAM's strong adaptability.

Fig.\;\ref{fig:fig6_tsne} further shows the visualization of objects of different classes in the feature  space learned with and without the explicit exploitation of inter-class correlation. As shown, with CAM introduced to capture inter-class correlation, object classes are better separated from each other, which affirms our motivation of leveraging inter-class correlation to reduce misclassification among similar classes.

\smallskip
\noindent\textbf{Number of Classes for Correlational Aggregation\;\;}
Meta-DETR receives a fixed number of support classes and simultaneously aggregates them with query features to capture the inter-class correlation among different support classes. Fig.\;\ref{fig:fig5_num_supp_class} investigates the impact of the number of support classes to aggregate at a time. As the number of support classes increases from 1 to 10, the lower-shot ($\leq$5) detection performance first improves and then drops, while 10-shot performance first saturates and then drops. This also validates the effectiveness of leveraging inter-class correlation under lower-shot settings. We conjecture the performance drop with a large number of support classes for correlational aggregation is due to the model's limited capacity to differentiate so many support classes at one go. Based on the results, we set our method's number of support classes as 5 in all other experiments unless otherwise stated.

\smallskip
\vspace{+0.5mm}
\noindent\textbf{Impact of Explicitly Modeling Background\;\;}
Table\;\ref{tab:ablation2} also validates the effectiveness of explicitly modeling a prototype and a task encoding for background, which allows our method to better handle the `no match' scenario where the query features do not match any of the support classes.

\section{Conclusion}

This paper presents a novel few-shot object detection framework, namely Meta-DETR. The proposed framework achieves \textit{(i)} pure image-level meta-learning, which lifts the constraints caused by novel classes' inaccurate region proposals, and \textit{(ii)} effective exploitation of inter-class correlation, which reduces misclassification and enhances generalization among similar or related classes. Despite its simplicity, our method achieves state-of-the-art performance over multiple few-shot object detection setups, outperforming prior works by large margins. We hope this work can offer good insights and inspire further researches in few-shot object detection.

\clearpage
{
\bibstyle{aaai}
\bibliography{aaai22.bib}
}

\clearpage

\section{Appendix}

This section provides more details of our proposed method and experimental results, which are omitted in the main paper due to space limitation. 

\subsection{Detailed Architecture of Meta-DETR}

The transformer encoder and decoder in the proposed Meta-DETR have similar setups as Deformable\;DETR\;\cite{DeformableDETR}. Concretely, both transformer encoder and decoder have 6 layers and adopt the multi-scale deformable attention module, with the proposed correlational aggregation module (CAM) counted as one encoder layer. The channel dimension $d$ is 256, and the intermediate dimension of fully-connected layers (FC) inside the transformer is 1024. The dropout probability is set to 0.1. The number of attention heads is 8. The number of object queries $N$ is 300.

\begin{figure}[t] 
\begin{center}
   \includegraphics[width=1.0\linewidth]{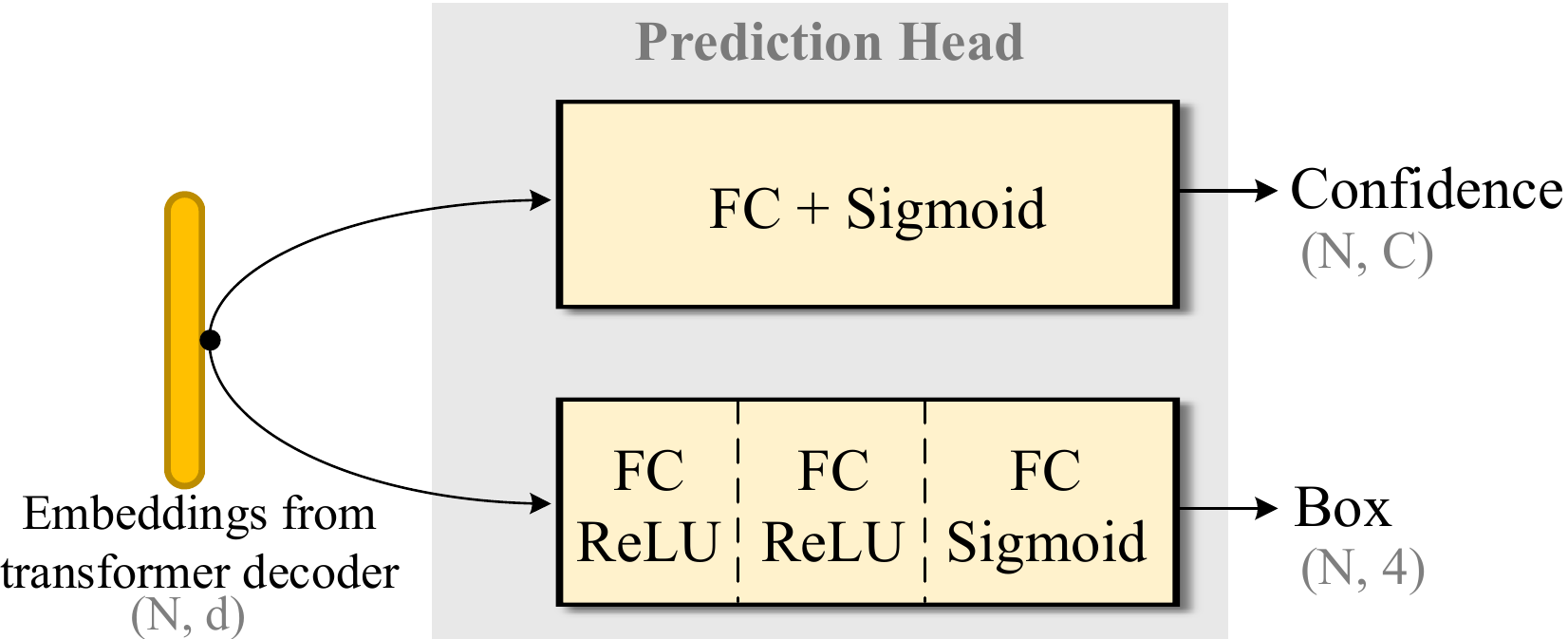}
\end{center}
\vspace*{-3.0mm}
   \caption{Illustration of the prediction head after the transformer encoder-decoder architecture to produce final predictions. It is shared for all the embeddings generated from the transformer decoder.}
\label{fig:pred_head}
\vspace*{-1.0mm}
\end{figure}

Fig.\;\ref{fig:pred_head} illustrates the prediction head that produces final predictions. The prediction head locates after the transformer encoder-decoder architecture, and is omitted for simplicity in Fig.\;\ref{fig:fig3_architecture} of the main paper. It consists of a 1-layer MLP for confidence prediction and a 3-layer MLP for box prediction. The prediction head is shared for all the embeddings generated from the transformer decoder.

\subsection{Training Objective of Meta-DETR}

Section 4.3 only provides a brief description of the training objective. Here, we provide a mathematically formulated description of the training objective in detail.

\subsubsection{Target Generation}
Let us denote the fixed number of object queries as $N$, which means Meta-DETR infers $N$ predictions within a single feed-forward process. Let us denote by $x_{\rm query}$ the query image, and $y \! = \! \left\{ y_i \right\}_{i=1}^{N}$ the ground truth set of objects within the query image, which is a set of size $N$. When $y_i$ indicates an object, $y_i \! = \! (c_i, b_i)$, where $c_i$ denotes the target class label and $b_i$ denotes the bounding box of the object. When $y_i$ indicates no object, $y_i \! = \! (\varnothing, \varnothing)$.

Meta-DETR dynamically conditions its detection targets on support classes and their mappings to the task encodings. As discussed in Section\;\ref{sec:meta-detr}, Meta-DETR predicts over $C$ support classes (\textit{i.e.}, target classes) simultaneously. The $C$ support classes are randomly sampled, denoted as $c_{\rm supp} \! = \! \left\{ s_i \right\}_{i=1}^{C}$. Besides, these support classes are further mapped to a set of task encodings. We denote the mapping function from the labels of support classes to the labels of task encodings as $\chi(\cdot)$. A specific case of $\chi(\cdot)$ can be formulated as:
\begin{equation}
\chi(s_i) = i \quad\quad i \in \{1,2, \cdots, C\}.
\end{equation}
Therefore, the detection targets of Meta-DETR can be formulated as:
\begin{equation}
y^{\prime} = \left\{ y^{\prime}_i \right\}_{i=1}^{N} = \left\{ (c^{\prime}_i, b^{\prime}_i) \right\}_{i=1}^{N} = \left\{ \psi(y_i, c_{\rm supp}) \right\}_{i=1}^{N},
\end{equation}
where $\psi(y_i, c_{\rm supp})$ acts to filter out irrelevant object annotations and to map the labels of target classes to the labels of the corresponding task encodings, which can be formulated as:
\begin{equation}
	\psi(y_i, c_{\rm supp})=\left\{
		\begin{aligned}
		&(\varnothing, \varnothing),& {\rm if}& \; \; y_i = (\varnothing, \varnothing)& \\ 
		&(\varnothing, \varnothing),& {\rm if}& \; \; c_i \notin c_{\rm supp}&.\\ 
		&(\chi(c_{i}), b_i),& {\rm if}& \; \; c_i \in c_{\rm supp}& \\ 
	\end{aligned}
	\right.
\end{equation}
Note that $y^{\prime}$ can completely consist of $(\varnothing, \varnothing)$ when there is no objects that belong to the provided support classes.

\subsubsection{Loss Function}
Assume the $N$ predictions for target class made by Meta-DETR are $\hat{y} \! = \! \left\{ \hat{y}_i \right\}_{i=1}^{N} \! = \! \big\{ (\hat{c}_{i}, \hat{b}_{i}) \big\}_{i=1}^{N}$. We adopt a pair-wise matching loss $\mathcal{L}_{\rm match}(y^{\prime}_i, \hat{y}_{\sigma(i)})$ to search for a bipartite matching between $\hat{y}$ and $y^{\prime}$ with the lowest cost:
\begin{equation} \label{eq:optimalassignment}
\hat{\sigma} = \mathop{\arg\min}_{\sigma} \sum\nolimits_{i=1}^{N}\mathcal{L}_{\rm match}(y^{\prime}_i, \hat{y}_{\sigma(i)}),
\end{equation}
where $\sigma$ denotes a permutation of $N$ elements, and $\hat{\sigma}$ denotes the optimal assignment between predictions and targets. Since the matching should consider both classification and localization, the matching loss is defined as:
\begin{equation} \label{eq:Lmatch}
\begin{split}
\mathcal{L}_{\rm match}(y^{\prime}_i, \hat{y}_{\sigma(i)}) = &\mathbbm{1}_{\left\{ c^{\prime}_i \neq \varnothing \right\}} \mathcal{L}_{\rm cls}(c^{\prime}_i, \hat{c}_{\sigma(i)}) \, + \\&\mathbbm{1}_{\left\{ c^{\prime}_i \neq \varnothing \right\}} \mathcal{L}_{\rm box}(b^{\prime}_i, \hat{b}_{\sigma(i)}) \;\;.
\end{split}
\end{equation}

With the optimal assignment $\hat{\sigma}$ obtained with Eq.\,\ref{eq:optimalassignment} and Eq.\,\ref{eq:Lmatch}, we optimize the network using the following loss function:
\begin{equation} \label{eq:L}
\begin{split}
\mathcal{L}(y^{\prime}, \hat{y}) \! = \! \sum\limits_{i=1}^{N} \left[ \mathcal{L}_{\rm cls}(c^{\prime}_i, \hat{c}_{\hat{\sigma}(i)}) + \mathbbm{1}_{\left\{ c^{\prime}_i \neq \varnothing \right\}} \mathcal{L}_{\rm box}(b^{\prime}_i, \hat{b}_{\hat{\sigma}(i)}) \right],
\end{split}
\end{equation}
where we adopt sigmoid focal loss\;\cite{focalloss} for $\mathcal{L}_{\rm cls}$ and adopt a linear combination of $\ell1$ loss and GIoU loss\,\cite{giouloss} for $\mathcal{L}_{\rm box}$. Similar to Deformable DETR\;\cite{DeformableDETR}, $\mathcal{L}(y^{\prime}, \hat{y})$ is applied to every layer of the transformer decoder.

Following Meta R-CNN\;\cite{metarcnn}, we additionally introduce a cosine similarity cross-entropy loss\;\cite{CloserFewshotClassification} to classify the class prototypes obtained by our designed CAM. It encourages prototypes of different classes to be distinguished from each other.

\subsection{Additional Comparison with State of the Art}

We also present results taking base classes into consideration in Table\;\ref{tab:Performance_VOC1_basenovel}. While achieving good performance for novel classes with limited training samples, Meta-DETR can still detect objects of base classes with competitive performance. TFA\,\cite{fsdet} produces outstanding performance for base classes since it works more like conventional detectors with fine-tuning, thus having relatively constrained capacity in generalizing on novel classes. We also wish to highlight that our proposed Meta-DETR achieves the best base-class and novel-class performance of all the compared meta-learning-based methods.

\begin{table}[t]
\begin{center}
\centering
\setlength{\tabcolsep}{2.688pt}
\resizebox{0.474\textwidth}{!}{
\begin{tabular}[t]{l|cccc|cccc}
\toprule[1.0pt]

&\multicolumn{4}{c|}{Base Classes} & \multicolumn{4}{c}{Novel Classes}\\

Method $\backslash$ Shot & 1 & 3 & 5 & 10 & 1 & 3 & 5 & 10 \\

\midrule[0.68pt]

Meta-YOLO\,\cite{FewshotReweighting} & 66.4 & 64.8 & 63.4 & 63.6 & 14.8 & 26.7 & 33.9 & 47.2 \\

FsDetView\,\cite{FSDetView}\,$\S$ & 64.2 & 69.4 & 69.8 & 71.1 & 24.2 & 42.2 & 49.1 & 57.4 \\

TFA\,w/\,cos\,\cite{fsdet}\,$\S$ & \textbf{77.6} & \textbf{77.3} & \textbf{77.4} & \textbf{77.5} & 25.3 & 42.1 & 47.9 & 52.9\\

MPSR\,\cite{MPSR}\,$\S$ & 60.6 & 65.9 & 68.2 & 69.8 & 34.7 & 46.1 & 49.4 & 56.7 \\

FSCE\,\cite{fsce}\,$\S$ & 75.5 & 73.7 & 75.0 & 75.2 & 32.9 & 46.8 & 52.9 & 59.7 \\

\rowcolor{black!6} Meta-DETR\,(Ours)\,$\S$ & 67.2 & 70.0 & 73.0 & 73.5 & \textbf{35.1} & \textbf{53.2} & \textbf{57.4} & \textbf{62.0} \\

\bottomrule[1.0pt]

\end{tabular}
}
\end{center}
\vspace*{-2.0mm}
\caption{Few-shot detection performance (mAP@0.5) for both base and novel classes on the first split of Pascal VOC. $\S$ indicates results averaged on multiple runs.}
\label{tab:Performance_VOC1_basenovel}
\vspace*{-0.5mm}
\end{table}

\subsection{Additional Ablation Study}

\subsubsection{Early Aggregation \textit{vs.} Late Aggregation}
We conduct experiments to study the location of our designed correlational aggregation module (CAM) to place. As shown in Table\;\ref{tab:ablation4}, it is preferable to place CAM at the beginning of the transformer encoder, which implies the importance of learning a deep class-agnostic predictor.

\subsection{Qualitative Results}

\begin{figure*}[t!] 
\begin{center}
   \includegraphics[width=1.0\linewidth]{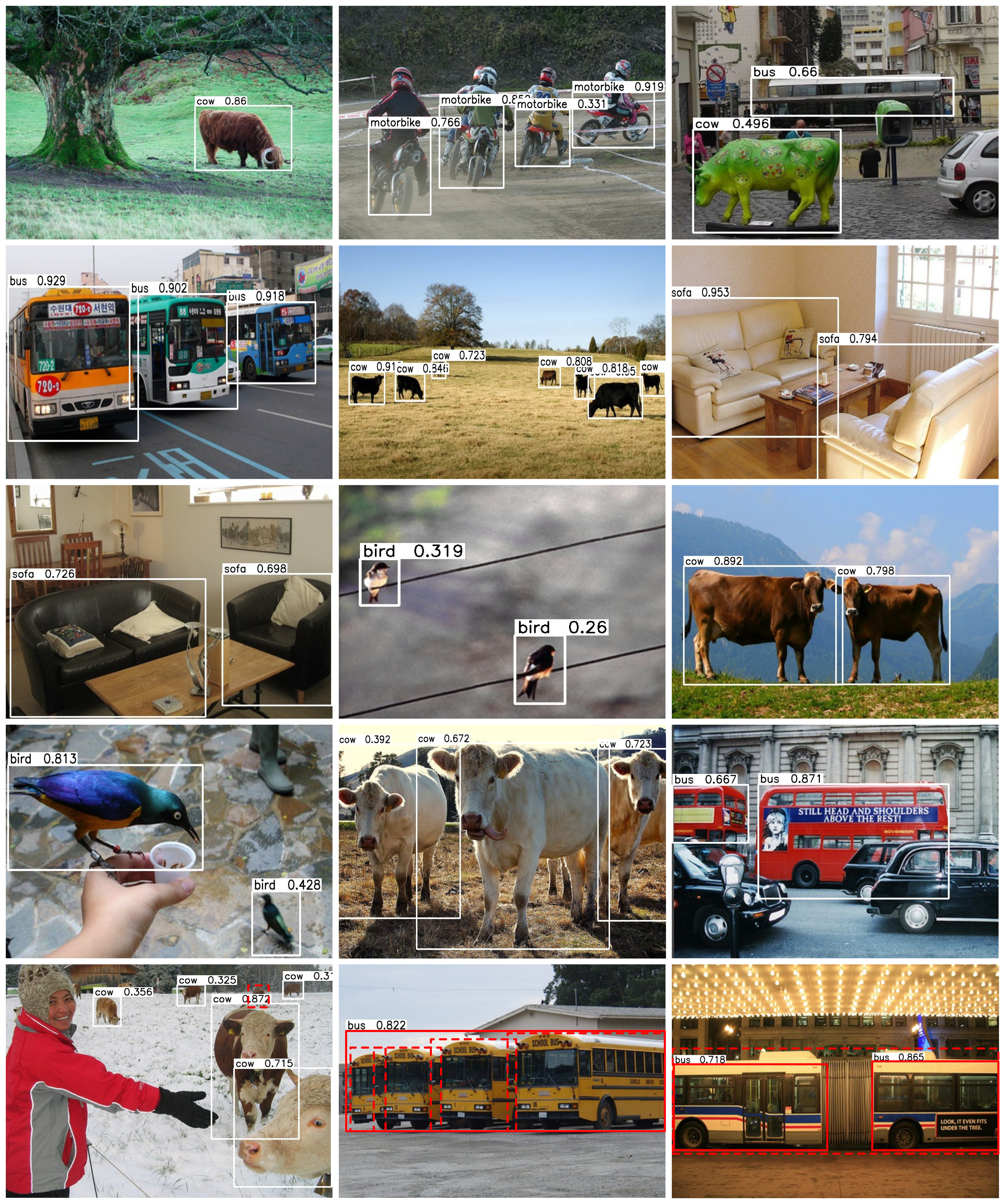}
\end{center}
\vspace*{-2.0mm}
   \caption{Visualization of Meta-DETR's 10-shot object detection results on Pascal VOC class split 1. Novel classes include bird, bus, cow, motorbike, and sofa. For simplicity, only results of novel classes are illustrated. White boxes indicate correct detections. Red solid boxes indicate false positives. Red dashed boxes indicate false negatives.}
\label{fig:supp_results_voc1_10shot}
\end{figure*}

\begin{figure*}[t!] 
\begin{center}
   \includegraphics[width=1.0\linewidth]{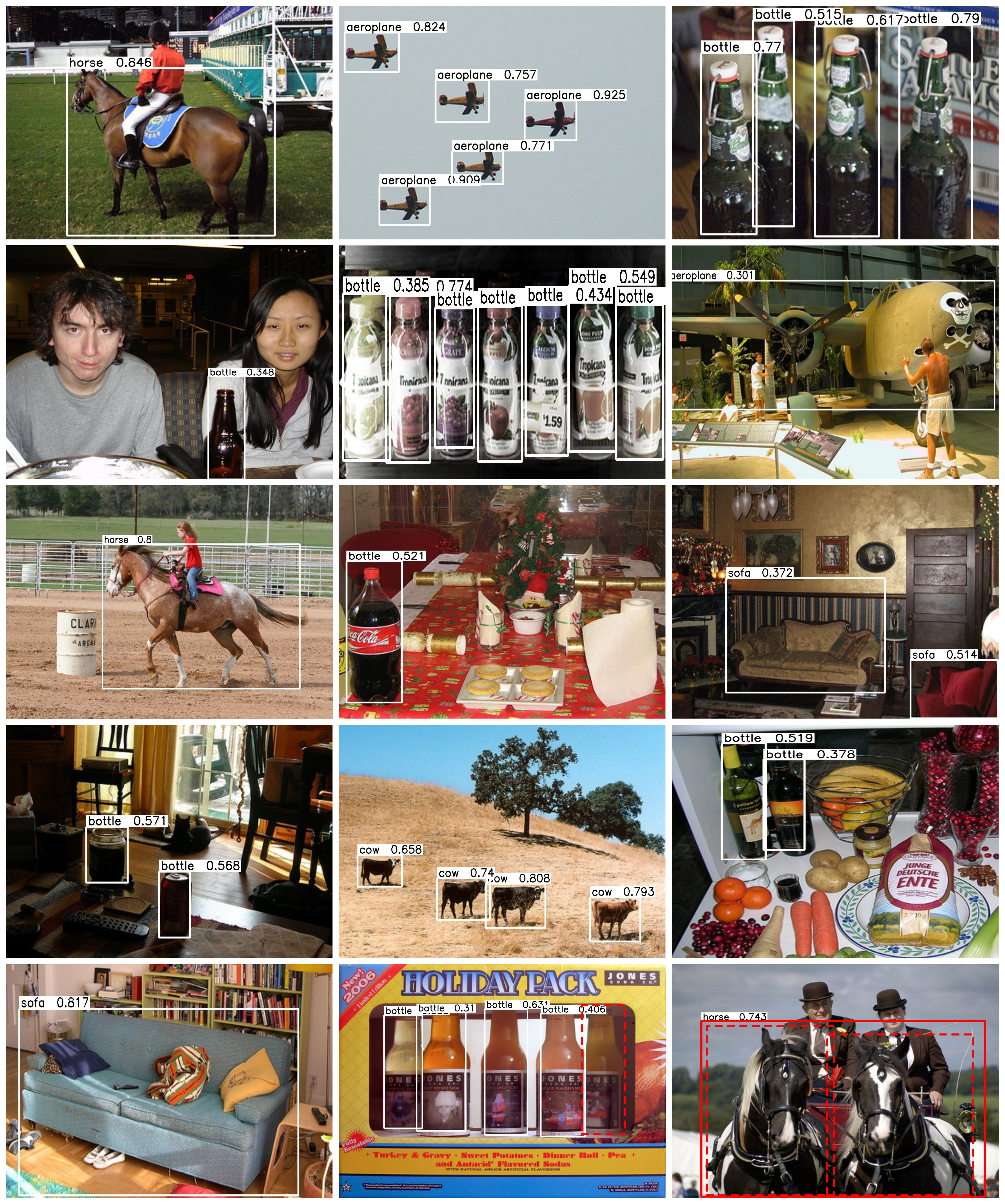}
\end{center}
\vspace*{-2.0mm}
   \caption{Visualization of Meta-DETR's 10-shot object detection results on Pascal VOC class split 2. Novel classes include aeroplane, bottle, cow, horse, and sofa. For simplicity, only results of novel classes are illustrated. White boxes indicate correct detections. Red solid boxes indicate false positives. Red dashed boxes indicate false negatives.}
\label{fig:supp_results_voc2_10shot}
\end{figure*}

\begin{figure*}[t!] 
\begin{center}
   \includegraphics[width=1.0\linewidth]{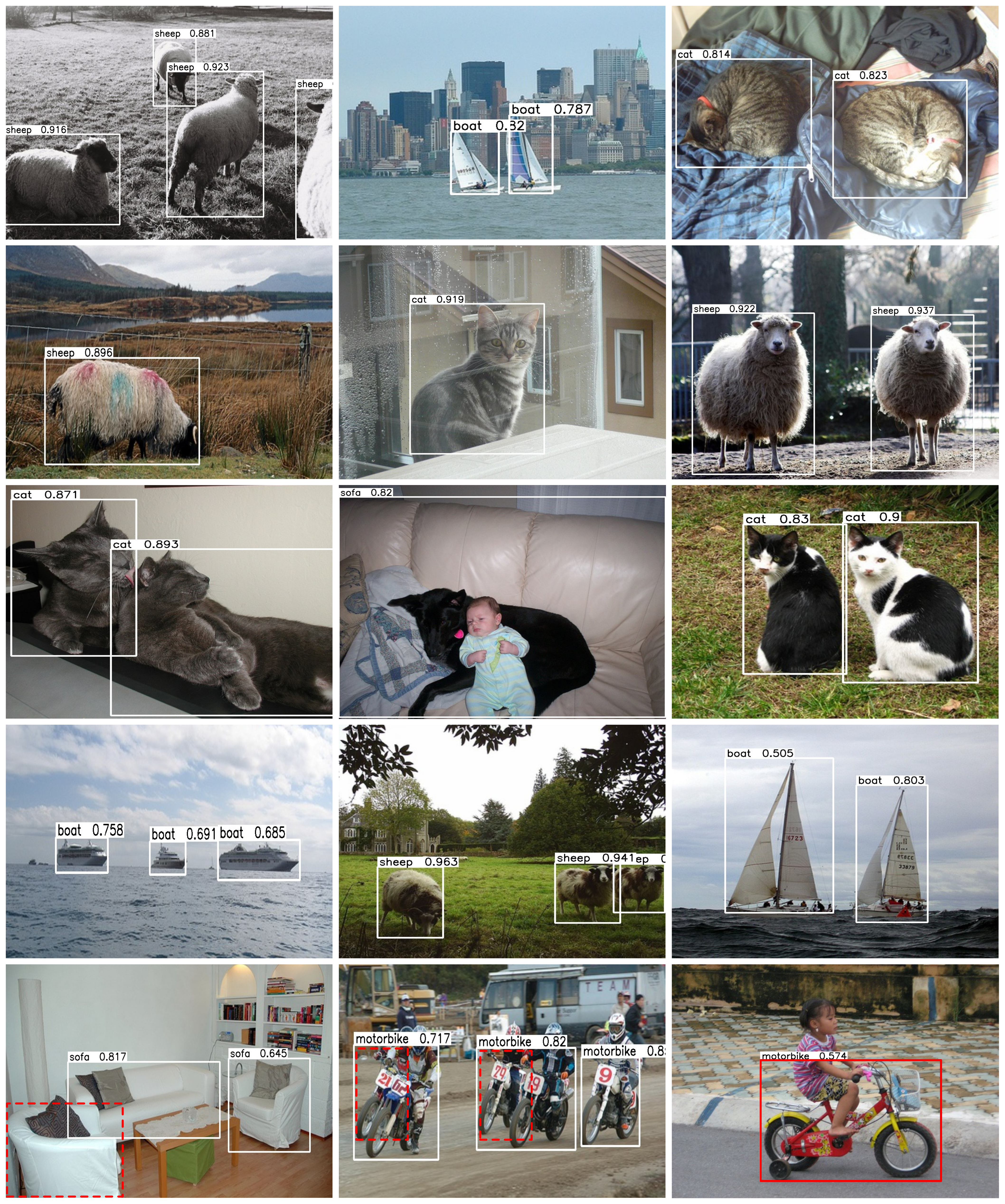}
\end{center}
\vspace*{-2.0mm}
   \caption{Visualization of Meta-DETR's 10-shot object detection results on Pascal VOC class split 3. Novel classes include boat, cat, motorbike, sheep, and sofa. For simplicity, only results of novel classes are illustrated. White boxes indicate correct detections. Red solid boxes indicate false positives. Red dashed boxes indicate false negatives.}
\label{fig:supp_results_voc3_10shot}
\end{figure*}

\begin{figure*}[t!] 
\begin{center}
   \includegraphics[width=1.0\linewidth]{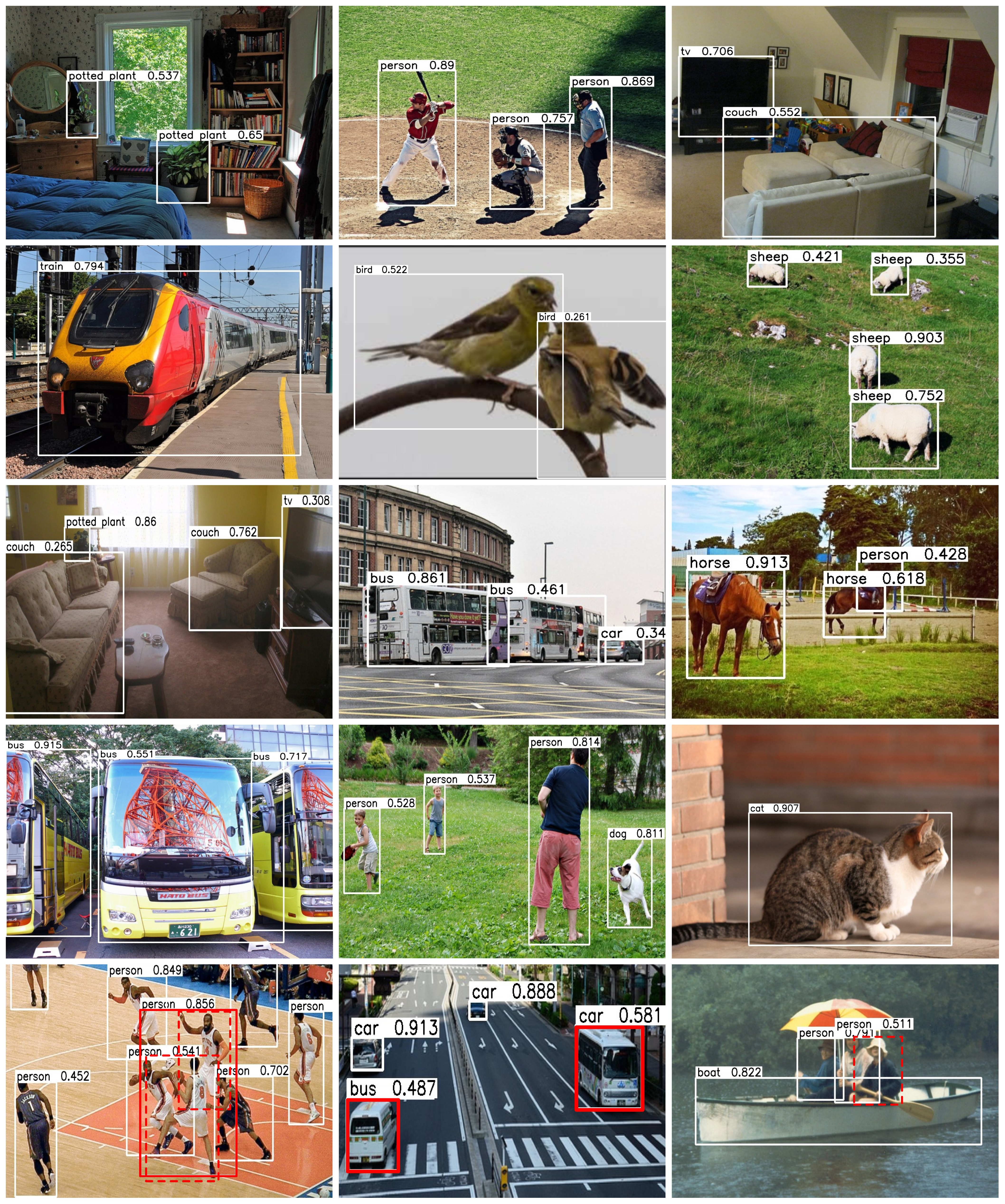}
\end{center}
\vspace*{-2.0mm}
   \caption{Visualization of Meta-DETR's 10-shot object detection results on MS COCO. Novel classes include person, bicycle, car, motorcycle, airplane, bus, train, boat, bird, cat, dog, horse, sheep, cow, bottle, chair, couch, potted plant, dining table, and tv. For simplicity, only results of novel classes are illustrated. White boxes indicate correct detections. Red solid boxes indicate false positives. Red dashed boxes indicate false negatives.}
\label{fig:supp_results_coco_10shot}
\end{figure*}

We provide multiple qualitative visualizations of Meta-DETR's few-shot detection results in Figs.\;\ref{fig:supp_results_voc1_10shot}-\ref{fig:supp_results_coco_10shot}, which give a straightforward illustration of the performance of our method. Note that only detection results of novel classes are presented, as the major focus is to detect objects of novel classes. In addition, we only show results with confidence scores higher than 0.25. White boxes indicate correct detections, red solid boxes indicate false positives, and red dashed boxes indicate false negatives. It can be observed that the proposed Meta-DETR is able to detect novel objects at a satisfactory performance even with scarce training samples.

In addition, we also provide a demo video attached in the supplementary materials, which consists of several short clips with Meta-DETR's predictions on novel classes as a reference.

\begin{table}[t]
\begin{center}
\centering
\setlength{\tabcolsep}{6.666pt}
\resizebox{0.398\textwidth}{!}{
\begin{tabular}[t]{c|ccccc}
\toprule[1.0pt]

CAM's Location & \multicolumn{5}{c}{Novel mAP@0.5} \\

@ Encoder Layer & 1 & 2 & 3 & 5 & 10 \\

\midrule[0.56pt]

1 & \textbf{35.1} & \textbf{49.0} & \textbf{53.2} & \textbf{57.4} & \textbf{62.0} \\

3 &  27.1 & 42.9 & 50.6 & 54.0 & 59.2 \\

6 & 15.2 & 31.5 & 37.7 & 50.3 & 53.4 \\

\bottomrule[1.0pt]

\end{tabular}
}
\end{center}
\vspace*{-1.5mm}
\caption{Ablation study on the location of our designed CAM. Results are averaged over 10 repeated runs on the first class split of Pascal VOC.}
\label{tab:ablation4}
\vspace*{-0.0mm}
\end{table}

\end{document}